%% file: main.tex
\documentclass[10pt,twocolumn,letterpaper]{article}

\usepackage[final]{cvpr} 

\input{preamble} 
\definecolor{cvprblue}{rgb}{0.21,0.49,0.74}
\usepackage[pagebackref,breaklinks,colorlinks,allcolors=cvprblue]{hyperref}
\hypersetup{
  pdfauthor={Miriam Horovicz, Yacov Hel-Or, Yael Moses},
  pdftitle={Diamonds in the Sky: Pareidolic Animals in Clouds}
}

\usepackage[accsupp]{axessibility}  
\usepackage{amsmath,amssymb}
\usepackage{graphicx}
\usepackage{multirow}
\usepackage{algorithm}
\usepackage{algpseudocode}
\usepackage{listings}
\usepackage{color}
\usepackage{float}
\usepackage{booktabs}
\usepackage{xcolor}

\newcommand{\comment}[1]{}                     

\def\paperID{14158}
\def\confName{CVPR}
\def\confYear{2026}

\title{Diamonds in the Sky: Pareidolic Animals in Clouds}
\author{Miriam Horovicz \qquad Yacov Hel-Or \qquad Yael Moses\\
Reichman University, Israel\\
{\tt\small miriam.horovicz@post.runi.ac.il, toky@runi.ac.il, yael@runi.ac.il}}

\begin{document}
\maketitle

\begin{abstract}
People often see animal shapes in clouds, a phenomenon known as pareidolia.
We propose an AI-based method that aims to predict which animals people are likely to perceive in clouds, even though state-of-the-art recognition methods typically fail to detect such animals.
Additionally, we introduce a method to assist individuals in perceiving specific pareidolic animals, even if they did not recognize them initially.

Our approach uses a diffusion model to transform cloud segments into an animal shape that visually resemble the original cloud.
This diffusion technique is inspired by the observation that the diffusion process succeeds only when the target animal resembles the shape of the cloud, and that subtle visual hints often suffice to help individuals recognize specific pareidolic animals. A generated image, successfully derived from the diffusion model, is then used to predict the pareidolic animal. Additionally, a short morphing video transitioning from the generated image back to the original cloud segment is employed to further enhance the human's perception of the pareidolic animals.
\end{abstract}

\section{Introduction}

\begin{figure}[!t]
    \centering
    \begin{tabular}{c|cc}
         \includegraphics[width=0.14\textwidth]{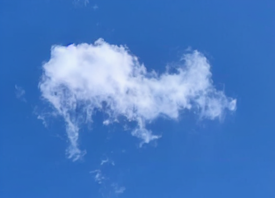} & \includegraphics[width=0.14\textwidth]{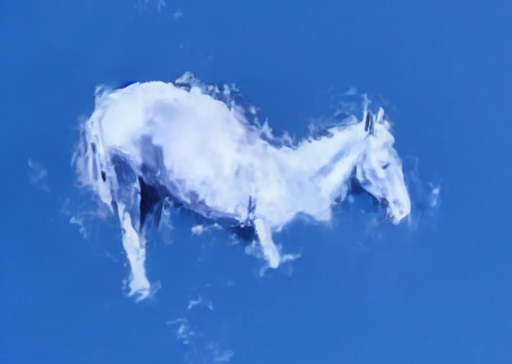} & \includegraphics[width=0.14\textwidth]{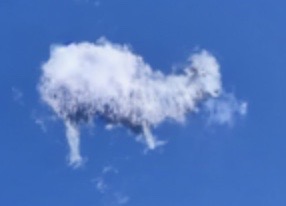}\\
         Original & Horse & Sheep 
             \end{tabular}
       \caption{
       Our method predicts pareidolic sheep and horse in the original cloud image, aligning with the perception of 60\% of participants in our survey. In contrast, SOTA recognition systems (OneFormer \& CLIP) failed to recognize any animal in the original image.}
    \label{fig:results}
\end{figure}

A child excitedly points at the cloudy sky during recess: “Look, a horse!” (see, Figure~\ref{fig:results}).
The child's classmates gather, some see the horse, others a sheep or a zebra, while a few perceive only the original cloud. Their perception may shift if a hint is given, such as pointing out where the horse’s head is. This phenomenon, known as pareidolia, is common and highlights humanity’s innate ability to recognize familiar shapes in random cloud formations.

In this work, we address two computational aspects of pareidolia in cloud images: (i) predicting which animals are likely to be perceived in a given cloud image, and (ii) enhancing the human ability to perceive specific pareidolic animals, if possible, even if they were not initially perceived.

A straightforward approach for predicting pareidolic animals is to use a state-of-the-art (SOTA) object detector. 
Object detectors (e.g., OneFormer \cite{jain2022oneformer}) or zero-shot classification (e.g., CLIP \cite{radford2021learning}) typically fail to detect the pareidolic animal as the object does not actually appear in the image, and the shape of the cloud lacks clarity and context.
In contrast, humans can still perceive pareidolic animals. 
\comment{To overcome these limitations in detecting pareidolic {\em faces} in natural images, Kumar et al.~\cite{faiman2024seeing} fine-tuned a face detection network on diverse datasets, improving its ability to recognize face-like patterns.} 
A previous attempt to predict pareidolic {\em faces} in natural images (\cite{hamilton2024seeing}) fine-tuned a face detection model on diverse datasets, to improve its ability to recognize patterns similar to those of  faces. In contrast, our method does not require additional datasets or further training and can work in a zero-shot manner. 

Our method is inspired by the observation that a subtle visual hint is often enough to help an individual recognize a specific pareidolic animal. The core idea is to apply slight modifications to a segmented cloud, extracted from a sky image, ensuring that the modified image (i) closely resembles the original segmented cloud, and (ii) is recognizable by a detection model. 
The modified image is then used for the prediction of the pareidolic animal and a short {\it guided video}, created by morphing between the generated image and the original cloud segment, is used to improve human perception of the pareidolic animal.

We implement this concept by diffusing a segmented cloud image toward an animal shape within the same segment using a proposed {\em Masked Delta Denoising Score} (MDDS) method. In a post-processing step, a modified cloud image is selected if it satisfies both recognizability and similarity criteria.

To evaluate our approach, we collected a cloud dataset and conducted an extensive user perception study. The results demonstrate a strong correlation between our predictions and human perception. Furthermore, the guided videos shown to participants significantly enhance their perception of the pareidolic animals.

Our main contributions are:
(1) Introducing the new task of pareidolic animal detection in clouds, which addresses the computational challenge of identifying animal shapes in ambiguous formations;
(2) Utilizing a zero-shot image modification rather than training specialized object detectors, enabling our approach to function without additional training or fine-tuning;
(3) Introducing the {\em Masked Delta Denoising Score} (MDDS), a region-constrained text-to-image method that can diffuse cloud segments while preserving the background elements; and
(4) Implementing a comprehensive evaluation methodology through a perceptual user study, validating both the predictive power of our approach and its effectiveness in enhancing pareidolic animal perception in clouds.

In a wider scope, this research opens new questions about pareidolic recognition, shifting from traditional object detection toward modification of abstract elements into recognizable forms. This work reveals a fundamental gap in current vision systems: while humans readily perceive animal shapes in clouds, state-of-the-art detectors fail completely. The method enables applications in virtual and augmented reality, such as procedural generation of natural environments, as well as computational tools for cognitive scientists studying pattern perception. Our approach demonstrates how diffusion models can be adapted to model human-like interpretive perception, opening new directions for perceptually-aligned vision systems.

\section{Previous Works}
Pareidolia refers to the human tendency to perceive meaningful patterns, such as faces or animals, in ambiguous stimuli. The neurological basis for pareidolia has been studied by Liu et al. \cite{liu2014seeing} and Wardle et al. \cite{wardle2022face}, highlighting how the human brain is specifically tuned to recognize faces even in highly abstract patterns.

Recent work by Hamilton et al. \cite{hamilton2024seeing} identified a significant gap between human perception and machine detection of pareidolic faces, introducing a dataset of 5,000 images containing face-like patterns in various objects. Gupta and Dobs \cite{gupta2025human} demonstrated that deep neural networks trained on both face identification and object categorization can develop human-like pareidolic perception, suggesting that pareidolia may result from the visual system's optimization.

Computational approaches to pareidolia have been explored. Song et al. \cite{song2021everything} proposed a novel approach with ``Pareidolia Face Reenactment'', animating static illusory faces to move in tandem with human facial expressions in videos.

The field of object detection has made significant strides with deep learning and transformer-based models such as OneFormer~\cite{jain2022oneformer}, DETR~\cite{carion2020end}, and Mask2Former~\cite{cheng2022masked}. Vision-language models like CLIP~\cite{radford2021learning} and OWL-ViT~\cite{minderer2022simple} have further advanced the capabilities of zero-shot object recognition. However, these models excel at detecting well-defined objects with clear boundaries and relevant context. When applying these detectors to cloud images, even state-of-the-art detectors fail to recognize pareidolic animals. This limitation highlights the lack of imaginative capacity in machine perception and illustrates the gap between conventional object detectors and human perception.

A recent work~\cite{hamilton2024seeing} aimed to detect face structures in abstract forms. Their solution involved fine-tuning a pre-trained object detection network on diverse and abstract datasets. 
While faces benefit from strong, meaningful features that make them easier to detect, animal shapes in clouds present greater variability and ambiguity. 

Our approach extends the concept of triggering pareidolia beyond faces to encompass a broader range of animal forms. Instead of fine-tuning networks with additional datasets, our method enhances and reveals animals in clouds through controlled image modifications. We leverage diffusion models and require neither additional training nor laborious data collection. 


\section{Proposed Method}
The human ability to imagine an animal shape in abstract forms is emulated in our work by applying subtle modifications to the original image that gradually reveal animal features, if possible. We demonstrate that object detectors can identify a pareidolic animal in the modified image, even when they fail to do so in the original cloud formation. In our pipeline, we use OneFormer model~\cite{jain2022oneformer} for object detection and segmentation, which is limited in the set of relevant objects it can recognize. Future advancements in object segmentation and detection methods are expected to further improve and extend our approach, enabling it to handle a wider range of pareidolic animals.

\begin{figure*}[!t]
    \centering
    \includegraphics[width=0.9\textwidth]{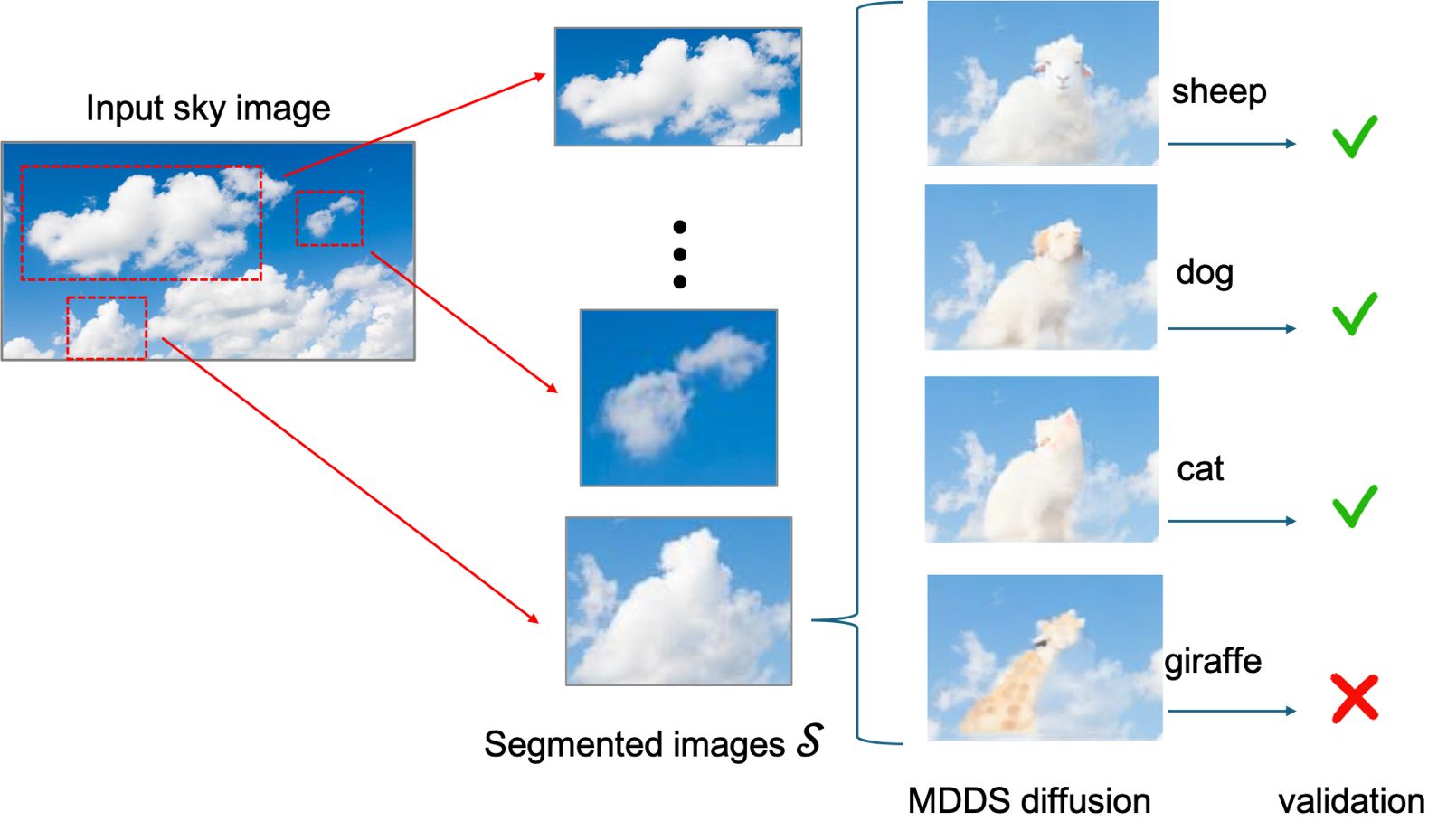}
    \caption{
    Pipeline overview. The input sky image is segmented into separate cloud segments using SAM, and the top-$k$ cloud segments are selected. Each segment is diffused toward 9 different animals using MDDS, generating modified segments. These are screened for recognizability and  similarity. Only modified segments that pass both criteria are accepted as valid segments.
    }
    \label{fig:pipeline}
\end{figure*}


A sky image, $I$, is segmented  into a set, $\mathbb S$, of $K$ cloud segments,
where a binary mask $mask(S)$ for $S\in \mathbb S$ indicates the spatial location of $S$ in the image plane. 
Each  cloud segment is analyzed with respect to  a set of animals,~$\mathcal A$, that might be perceived in the cloud segment. That is, for each selected segment~$S\in \mathbb S$ and animal $a\in \mathcal{A}$ we aim to address two tasks: Task~1 – predicting whether  animal $a$ will be pareidolically perceived in the cloud segment~$S$, and Task~2 – enhancing an observer's perception of the pareidolic animal $a$ if this animal can be perceived in $S$. 

The key part of our solution for both tasks is to slightly modify the segmented cloud $S$ toward a segment $\hat M^a$, that should resemble both the animal $a$ and the original segments~$S$.
Our method leverages the  text-to-image diffusion models (e.g., \cite{rombach2022high,ramesh2022hierarchical,saharia2022photorealistic}), incorporating a novel adaptation, in order to generate $\hat M^a$. The segment $\hat M^a$ is obtained by interpolating between a diffused image $M^a$ and the original image, ensuring that it passes the validation test: (i) the animal $a$ can be recognized in $\hat M^a$ by an object-detection model, and (ii) the segmented animal $a$ within $\hat M^a$ is similar in shape and color to the original cloud segment $S$ (see Sec.~\ref{sec:filtering}).
\comment{The  segment $\hat M^a$
is considered a {\em valid segment} if two conditions hold: (i) the animal $a$ can be recognized in $\hat {M}^a$ by an object detection model, and (ii) the segmented animal $a$ inside $\hat M^a$ is similar in shape and color to the original cloud segment ${S}$ (see Sec. \ref{sec:filtering}).} It is also used to enhance the observer's perception by producing a morphing video from the original segment 
$S$ towards $\hat {M}^a$ and back.
The entire process of the proposed approach is illustrated in Figure~\ref{fig:pipeline} and summarized in the pseudo-code in Algorithm~\ref{alg:pareidolia}.  We next 
elaborate on each part of this process.

\begin{algorithm}[ht]
\caption{Pareidolic Animal Prediction and Enhancement}
\label{alg:pareidolia}
\begin{small} 
\begin{algorithmic}[1]
\Statex \textbf{Input:} Cloud image $\mathcal{I}$, a list of animals $\mathcal{A}=\{a_1..a_\ell\}$
\Statex \textbf{Output:}
(i) A set of {\it valid segments}, each associated with a set of its valid pareidolic  animal. 
(ii)~A set of {\it morphing videos}:  for each valid segment and its animals.
\State Segment image $\mathcal{I}$ into $\{S_j\}_{j=1}^K$ using SAM
\For{each cloud segment $S \in \{S_1 ... S_K\}$}
    \For{each animal $a \in \mathcal{A}$}
        \State ${M}^a \gets \text{TextGuidedDiffusion}(S, a)$
        \If{animal $a$ is detected in ${M}^a$}
        \State Generate an interpolated segment $\hat M^a$ (see text)
        \If { ${\hat M}^a$ is similar to $S$}
            \State Add $(S,a)$ to the set of {\it valid-segments}
            \State Generate morphing video between $S$ and ${\hat M}^a$,
            \State ~~~~~~~~~~and add to the set of {\it morphing videos}
            \EndIf
        \EndIf
    \EndFor
\EndFor
\State \Return {\it valid segments} and {\it morphing videos}
\end{algorithmic}
\end{small}
\end{algorithm}
\begin{figure*}[!t]
\begin{center}
    \begin{tabular}{ccccc}
       \includegraphics[width=0.15\textwidth]{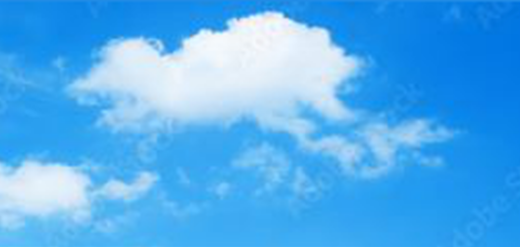} &          
       \includegraphics[width=0.15\textwidth]{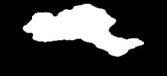} &
       \includegraphics[width=0.15\textwidth]{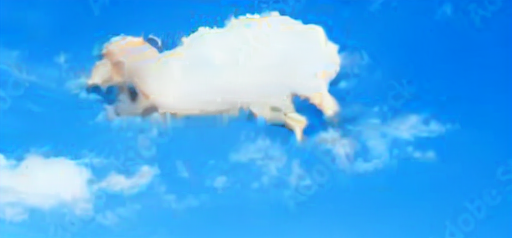} & 
       \includegraphics[width=0.15\textwidth]{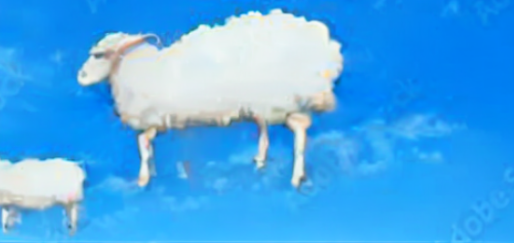} & 
       \includegraphics[width=0.15\textwidth]{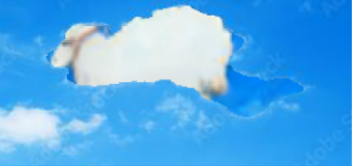} \\
       (a) Original  & (b) Mask & (c) MDDS  &       
             (d) DDS & (e) DDS - masked \\
                        \end{tabular}
                        \end{center}
    \caption{(a) An original image; (b) The mask; (c) Our MDDS-based diffusion toward a sheep using the cloud mask;
(d) DDS-based diffusion toward a sheep. Note, that 2 sheeps were generated and the modifications occur also outside of the mask. (e) The mask naïvely applied to the result in (d). The generated sheep lost its form.
    }
    \label{fig:dds_vs_mdds}
\end{figure*}
\subsection{Segmentation} 
In order to segment a cloud image into a set of cloud candidates, we employ the Segment Anything Model (SAM)~\cite{kirillov2023segment}. From all segments $\{S_j\}_{j=1..N}$, we select the top-$K$ largest cloud segments, which usually correspond to prominent cloud formations (typically $K=5$). 
Each selected segment is cropped using its minimal bounding box. 
It is then padded with the average sky color to preserve the sky context, resulting in a square 
$512\times 512$ input image. This image is then  passed to the diffusion step. 

Figure~\ref{fig:pipeline} illustrates the process using $K$ segments from a sky image. Each segment is treated independently throughout the pipeline.


\subsection{Test-guided Diffusing}
Next, we describe our diffusion-based method for modifying an image segment $S$ toward an animal $a \in \mathcal{A}$. The input to the diffusion model is composed of an animal name $a$, a padded image segment ${S}$, and its segmentation  $\mbox{\em mask}(S)$. Our goal is to modify only the masked area of $S$ while leaving the remaining image areas untouched. 

Text-to-image diffusion models (e.g., Stable Diffusion, DALL-E, and Imagen~\cite{ramesh2022hierarchical, saharia2022photorealistic, rombach2022high})  often face challenges in controlling the generated image modifications. Recent advances, such as the Delta Denoising Score (DDS) \cite{hertz2023delta}, have introduced a probabilistic framework for controllable image manipulation. DDS utilizes the Score Distillation Sampling loss \cite{poole2022dreamfusion} to perform subtle and localized edits.

The intuition behind DDS is to identify the differences between how an image would be represented under different text descriptions, such as `a horse' versus `a cloud'. By controlling these differences in multiple optimization steps, DDS preserves much of the original structure while gradually introducing features of the target concept. Technically, DDS works through an optimization process that minimizes the following loss function:

\begin{equation}
\begin{array}{ccl}
L^t_{\mathrm{DDS}} & = & \left\| \epsilon_\theta(z_t, y_{\text{target}}, t) - \epsilon_\theta(z_t, y_{\text{source}}, t) \right\|^2
\end{array},
\end{equation}
where $z_t$ is the noisy latent image at timestep $t$, $\epsilon_\theta$ is the noise prediction from the diffusion model, $y_{\text{target}}$ is the target prompt (e.g., `a horse'), and $y_{\text{source}}$ is the source prompt (e.g., `a cloud').

 However, when modifying an image segment using DDS, it may also affect regions outside the intended mask (see example in Figure~\ref{fig:dds_vs_mdds}(d)).
 For our cloud pareidolia task, this limitation is critical, as we need to modify only the segmented cloud, $S$, while preserving the surrounding sky. 
Naïvely applying $\mbox{\em mask}(S)$ to the DDS result often degrades both the natural appearance of the cloud and the animal form (see Figure~\ref{fig:dds_vs_mdds}(e)). Therefore, we propose  a masked version of DDS, as next described.

\subsubsection{Masked Delta Denoising Score (MDDS).}
\begin{figure*}[t]
    \centerline{\includegraphics[width=1\linewidth]{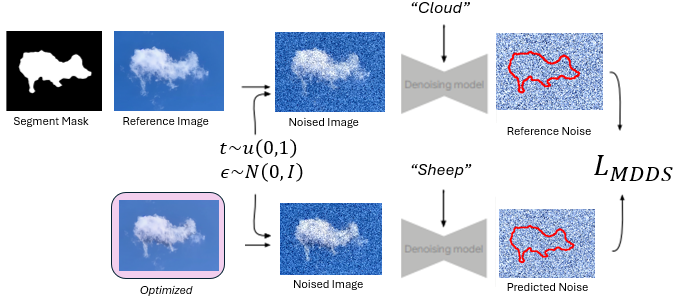}}
    \caption{Overview of our MDDS method.
Input: a reference image $S$ and a segment mask $mask(S)$. At each step, the same noise is added to the reference (first row) and the optimized (second row) images. For pixels inside the mask, we use the noise prediction from the target animal prompt (e.g., `sheep'), while for pixels outside the mask, we use the noise prediction from the source prompt (`cloud'). For visualization, the mask outline is marked in red on the predicted noise image. This selective combination creates a masked gradient that updates only the cloud region while preserving the background perfectly.
}
    \label{fig:mdds}
\end{figure*}

MDDS  extends DDS with spatial control through a binary mask that confines modifications to the segmented cloud region (see Figure~\ref{fig:mdds}). In  MDDS the  noise predictions is applied, at each step, based on spatial location:
\begin{equation}
L^t_{\mathrm{MDDS}}  =  \left\| \mbox{\em mask}(S) \odot (\epsilon_\theta(z_t, y_{\text{target}}, t) - \epsilon_\theta(z_t, y_{\text{source}}, t)) \right\|^2
\end{equation}
where $\mbox{\em mask}(S)$ is the binary mask (`1' inside the cloud region, `0' elsewhere), $\odot$ represents element-wise multiplication, and all other variables follow the same definitions as in Eq. (1). The noise in the  region outside of $\mbox{\em mask}(S)$ is taken to be the original one of $y_{\text{source}}$ (in our case `a cloud') and it does not affect the loss. Hence,  it is expected that denoising process outside the mask will not be affected. This process results in a diffused image $M^a$.

\subsection{Segment Validation}
\label{sec:filtering}
The diffused image,  $M^a$,  must be validated, as it may not contain a structure similar to the animal $a$, it may contain an animal structure that only occupies a small fraction of the segment~$S$ (see $a=$`giraffe' in Figure~\ref{fig:pipeline}),  or it may not resemble the original segment~$S$ at all. This usually happens when the original segment $S$ is not similar to the animal $a$ to begin with, hence we would like to discard this segment.
Thus, we first ensure that $M^a$ is recognizable, and then modify it to obtain $\hat M^a$ such that it preserves sufficient similarity to the original segment, when possible.
 
\paragraph{1. Recognizability:}
A recognition method is used to determine whether the target animal is recognizable in ${M}^a$. We use OneFormer \cite{jain2022oneformer} to apply both classification and instance segmentation on ${M}^a$. Images in which the animal $a$ is not recognized by OneFormer are discarded.
There are only 10 animal categories included in the  COCO~\cite{lin2014microsoft} dataset, on which OneFormer is pre-trained. Hence, we considered the set of animals  $\cal A $=\{bird, cat, dog, horse, sheep,  cow, elephant, bear, giraffe\}. The zebra from COCO is excluded due to its 
high visual similarity to the horse, which led to redundant predictions, and to reduce computational load during inference. If more advanced recognition and segmentation models become available, they could support the diffusion for a wider variety of objects from clouds.

\paragraph{2a. Spatial Similarity:}
To avoid cases where the generated animal occupies only a small fraction of ${M}^a$ as in Figure~\ref{fig:pipeline} (for `giraffe'), we use the  instance segmentation of OneFormer , $\mbox{\em mask}(M^a)$, of the detected animal $a$ in ${M}^a$.
Note that the instance segmentation, $\mbox{\em mask}(M^a)$,    
differs from the cloud segmentation, $\mbox{\em mask}(S)$, applied by SAM. 
Note, also that by construction $\mbox{\em mask}(M^a) \subseteq \mbox{\em mask}(S)$.

The $\mbox{\em mask}(M^a)$  should occupy a significant part of $\mbox{\em mask}(S)$.
 The spatial similarity between the two masks  is computed by intersection over unions (IoU):
\begin{equation}
IoU(S; M^a) = \frac{|\mbox{\em mask}(S) \cap \mbox{ \em mask} (M^a)|}{|\mbox{\em mask}(S)|}.
\end{equation}
To validate $M^a$, we require that $IoU(S; M^a) \geq 0.5$, otherwise ${M}^a$ is discarded (see the `giraffe' example in Figure~\ref{fig:pipeline}).

\begin{figure}[b]
    \flushright 
    \includegraphics[width=0.45\textwidth]{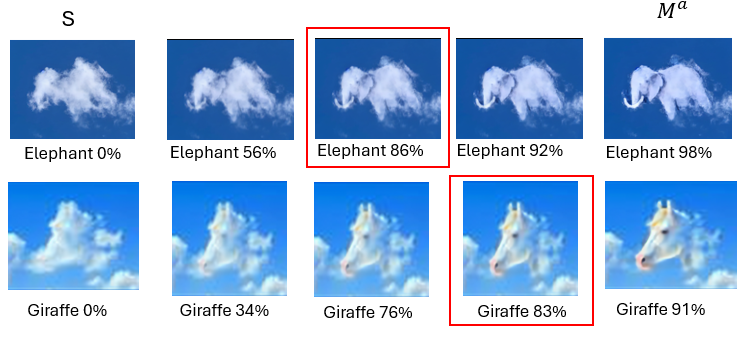}
    \caption{Interpolation results between the original image \(S\) and the  \(M^a\) for 2 different original cloud segments. Each image is annotated with its recognition confidence score (\%) towards the target animal. Red boxes indicate chosen images.}
    \label{fig:interpolation}
\end{figure}

\paragraph{2b. Photometric Similarity:} 
As pareidolic animals are seen in clouds only if the original cloud segment $S$ resembles the structure of the animal, we would like to find the minimal amount of modification applied to $S$ so that the animal $a$ is still recognizable. 
To do so,  a set of linear interpolations between $S$ and  ${M^a}$ is computed:
$$
{\hat M}^a(\alpha)= (1-\alpha) S + \alpha {M^a}~~,~~~~\mbox{where}~~\alpha \in [0..1]
$$
The recognition scores of OneFormer applied to ${\hat M}^a(\alpha)$ across various $\alpha$ values is computed. A typical score profile vs.~$\alpha$ shows raise in the recognition score as $\alpha$ increases. We chose the modified image ${\hat M^a}={\hat M^a}(\alpha_{rec})$ at the minimal $\alpha$ value (which is closer to the original segment) where the recognition score exceeds a threshold (in our implementation $0.8$) see Figure~\ref{fig:interpolation}. 

Assessing the degree of difference between $S$ and ${\hat M}^a$ 
requires robust metrics that align with human perception. We use the Structural Similarity Index (SSIM~\cite{wang2004image}), which compares local patterns of pixel intensities and is particularly effective at capturing structural similarity and perceptual quality.
To ensure that our method successfully preserves the original cloud structures, we choose only segments for which SSIM exceed a threshold (in our implementation $0.7$) between ${\hat M}^a$ and $S$.

\comment{While other metrics such as LPIPS~\cite{zhang2018unreasonable} and FID~\cite{heusel2017gans} are commonly used to evaluate generative image quality, our task differs: we are not assessing realism or diversity of generated images, but rather the preservation of the original cloud structure after transformation. In this context, SSIM is more appropriate, as it captures fine-grained structural changes and aligns better with our screening goal.

Therefore, we rely on SSIM in our screening pipeline to ensure that the A-images preserve key structural features of the original cloud segments. } 

\begin{figure*}[!t]
    \centering
     \includegraphics[width=1\textwidth]{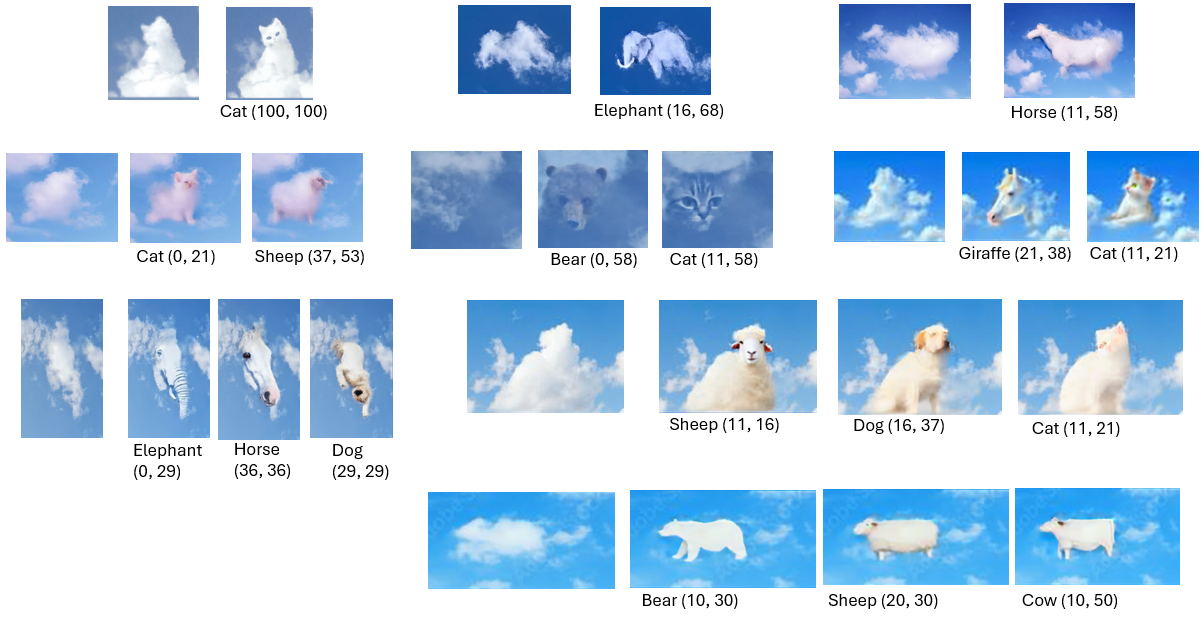}\\
    \caption{Examples of segment modifications, each corresponding to a different animal. The two numbers shown for each modified segment represent the number of participants in the user study who agreed with the animal to which the model assigned the original segment, before and after viewing the enhancement videos.
    \comment{Examples of multi-target generation. Each cloud can be transformed into multiple plausible animal shapes while preserving the original cloud structure. This approach captures the subjective nature of pareidolia, where different observers may perceive different animals in the same cloud. The two numbers shown for each image represent the accuracy of identification by participants in the user study: the first indicates the percentage who correctly identified the animal before seeing the video, and the second after.}
    }
    \label{fig:generation}
\end{figure*}

\subsection{Enhancement Video}
Not everyone necessarily perceives the same pareidolic animal in the same segment. However, subtle visual hints toward a specific animal can often strengthen or trigger pareidolia. To enhance a participant’s perception of the predicted animal, we generate a short \textit{enhancement video} by linearly interpolating between the original cloud segment $S$ and the resulting modified segment ${\hat M}^a$  over 30 steps, and then reversing it back. This creates a smooth morphing effect that subtly reinforces the intended animal shape while preserving the original cloud structure. 
Example enhancement videos are provided in the supplementary materials.

\section{Experiments}
Since there are no existing datasets with labeled cloud images of pareidolic animals, we collected a new dataset of 50 sky images\footnote{The dataset will be publicly available upon paper acceptance.}. Some of the images were captured by the authors, whereas others were obtained from publicly available datasets on the web, including Google Photos and Getty Images~\cite{googlephotos,gettyimages}.

We applied our method to the collected dataset of 50 sky images, which were segmented into 250 individual cloud regions. 
For each segment, we generated nine diffused images
resulting in a total of 2{,}250 
images (250 $\times$ 9).

During the validation phase, OneFormer~\cite{jain2022oneformer} successfully recognized an animal in 12\% of the diffused images (270 out of 2{,}250). Applying our validation procedure retained approximately 22\% of these recognized cases, resulting in 60 modified images used for prediction and perception enhancement. These modified images correspond to 23 unique cloud segments, as a single segment may be associated with more than one recognizable animal.
Example of segments and the generated images are shown in Figures~\ref{fig:results}–\ref{fig:dds_vs_mdds} and Figure~\ref{fig:generation}. More examples are available in the supplementary material.

\comment{Applying our validation retained approximately 22\% of the recognized cases (60 out of 270), yielding \red{60 validated} images overall. 
These \red{validated}  images correspond to 23 unique cloud segments, as it may happen that a segment is associated with more than one recognizable animal.

Examples of segments with validated predicted animals are shown in Figures~\ref{fig:results}–\ref{fig:dds_vs_mdds} and Figure~\ref{fig:generation}. More examples are available in the supplementary material.}

\subsection{Comparison to Recognition Systems}
We tested CLIP~\cite{radford2021learning} and OneFormer~\cite{jain2022oneformer} on the original, unmodified cloud segments. These models failed to detect any animals in any of these segments, including those identified and validated by our method. This is expected, as standard recognition models are trained to detect real, well-defined objects within clear contextual conditions - features that pareidolic animals in clouds typically lack. 
The abstract and ambiguous shapes of clouds pose a particular challenge for such models, helping to explain their failures and underscoring the need for methods like ours that account for human-like perception.

\subsection{Perceptual Study}
The two primary goals of our study are: (1) predicting which pareidolic animals humans perceive in the original cloud segments, and (2) enhancing human perception of specific pareidolic animals through subtle image modification.

\comment{We conducted a perceptual study to evaluate our method's
  effectiveness. Our dataset includes two types of segments: 1) {\em Validated (Positive)} segments: where the method successfully identified at least one animal in the cloud segment, and the enhanced version passed both recognizability and similarity tests, and 2) {\em None (Negative)}: the method found no animals in the cloud segment.
}

\paragraph{Survey.}
We collected 304 user responses across cloud segments. 
The study used two types of segments: {\em Positive} segments, where the method produced at least one valid $\hat M^a$, and {\em Negative} (None) segments, where no $\hat M^a$ could be generated.
Since each participant responded both, before and after viewing the enhancement, this yielded 608 total selections across 9 animal categories, resulting in 5,472 individual perception judgments.

Each participant was presented with a series of cloud segments.  
When prompted, participants were asked to indicate the animals they perceived in the segment or to skip the image if none were perceived. The animal selection was restricted to a fixed list, $\mathcal{A}$, comprising nine animals. The experiment followed a sequence of three steps for each segment:
\begin{itemize}
\item Step 1: The original cloud segment, $S$, was presented, and the participants were asked to choose a set of  animals, 
$A \subseteq \mathcal{A}$ 
they could perceive.
\item Step 2: One or more enhancement videos were displayed only for animals predicted by the  model for this segment, if any existed. Otherwise, an arbitrary video was shown. No response was required at this step.
\item Step 3: The original cloud segment, $S$, was presented  again, and the participants were asked  to choose again the set of animals, $A \subseteq \mathcal{A}$ they could perceive.
\end{itemize}
\vspace{0.5cm}
We next present and discuss  the obtained results.

\begin{figure}[!t]
    \centering
    \includegraphics[width=0.85\linewidth]{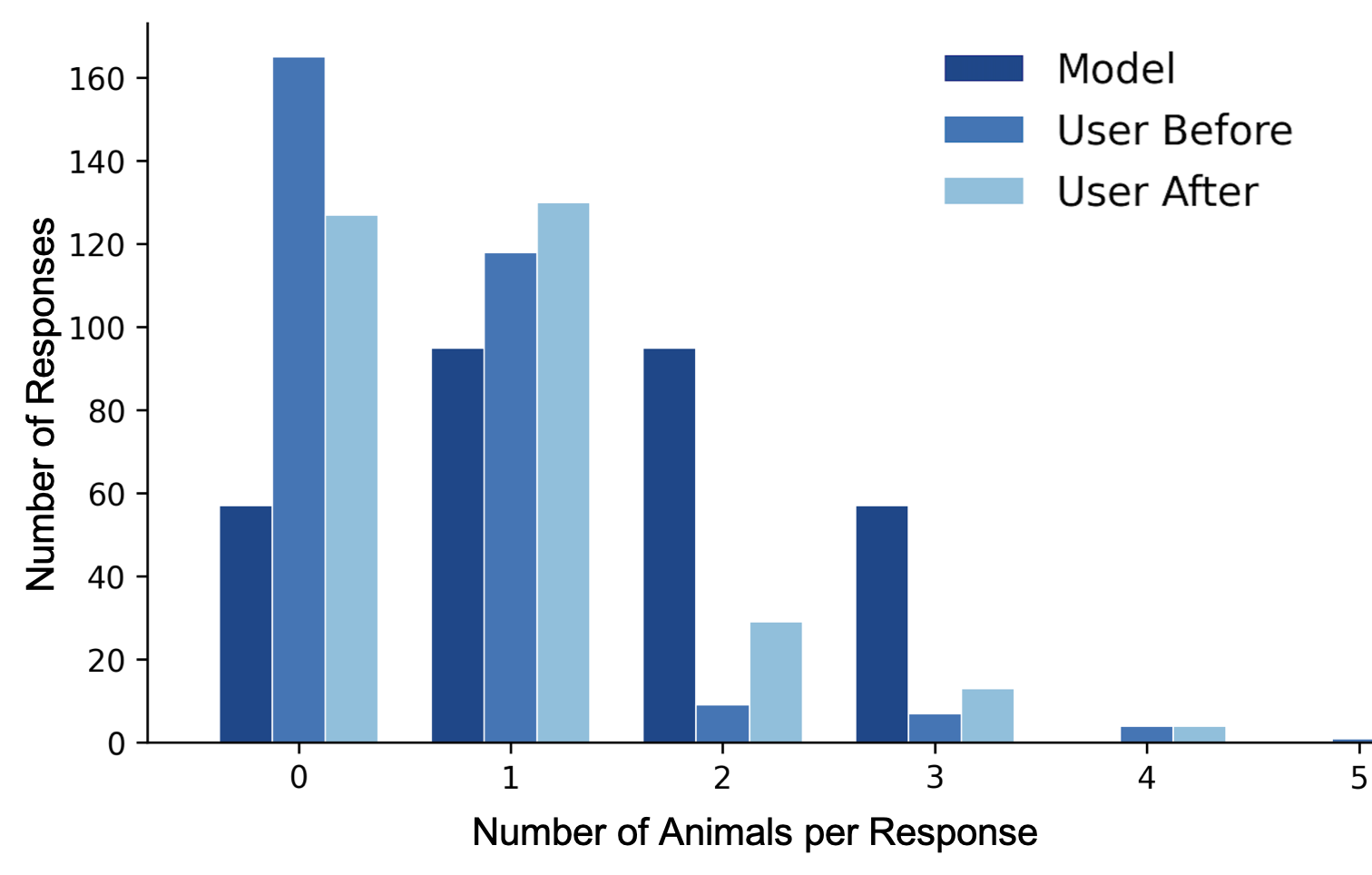}
    \caption{Distribution of the number of animals selected per response: comparison between model predictions, user responses before, and after seeing enhanced video.}
    \label{fig:histogram_response_counts}
\end{figure}

\paragraph{Distribution of Perceived Animals.}
We begin by analyzing the overall distribution of the number of animals identified per response. Figure~\ref{fig:histogram_response_counts} compares the number of animals  selected by participants (per segment) before and after viewing the enhancement videos, and those predicted by the model for the same segment.

Before viewing the enhancement, most participants identified  one animal or none. After enhancement, responses shifted toward one or two animals, suggesting improved clarity and increased confidence.

\paragraph{Prediction Accuracy.}
To evaluate how well the model’s predictions match human perception, we performed a per-animal, per-segment binary classification analysis using human responses as ground truth. For each segment, $S$, and animal category $a$, we compute a binary confusion matrix with the following outcomes: the model predicts $a$ and the participant also perceives $a$ (TP), the model predicts $a$ but the participant does not perceive it (FP), the model does not predict $a$ but the participant perceives it (FN), and neither the model nor the participant perceives $a$ (TN). We aggregate results across all 9 animals and responses for each segment yielding a single aggregated confusion matrix representing all 2,736 data points (304 responses × 9 animals). Table~\ref{tab:conf_matrix_after} presents these aggregated confusion matrices. 
The precision, recall, and F1 scores are summarized in Table~\ref{tab:baseline_comparison}.

\comment{To evaluate whether our model’s predicted animals align with human perception, we conducted a per-animal binary classification analysis using human responses as the ground truth.

For each of the 9 animal categories, we compute a confusion matrix comparing model predictions against human perception, where:
\begin{itemize}
\item \textbf{True Positive (TP):} Both human and model identify the animal as present.
\item \textbf{False Positive (FP):} Model predicts the animal, but human does not perceive it.
\item \textbf{False Negative (FN):} Human perceives the animal, but model did not predict it.
\item \textbf{True Negative (TN):} Both agree the animal is not present.
\end{itemize}

Using human responses as the reference (ground truth), we compute precision and recall. We aggregate results across all 9 animals by summing all confusion matrix values, yielding a single aggregated confusion matrix representing all 2,736 data points (304 responses × 9 animals). Table~\ref{tab:conf_matrix_after} presents these aggregated confusion matrices.}

\begin{table*}[!ht]
\centering
\vspace{0.3em}

\begin{minipage}{0.45\linewidth}
\centering
\textbf{Before enhancement}\\[0.3em]
\begin{tabular}{|l|c|c|}
\hline
 & Human: Yes & Human: No \\
\hline
Model: Yes & 49 & 407 \\ \hline 
Model: No & 129 & 2151 \\
\hline
\end{tabular}
\label{tab:conf_matrix_before}
\end{minipage}
\hfill
\begin{minipage}{0.45\linewidth}
\centering
\textbf{After enhancement}\\[0.3em]
\begin{tabular}{|l|c|c|}
\hline
 & Human: Yes & Human: No \\
\hline
Model: Yes & 138 & 318 \\
Model: No & 111 & 2169 \\
\hline
\end{tabular}
\end{minipage}
\caption{Aggregated confusion matrices \textbf{before} and \textbf{after} enhancement, summed across all segments, animals and participants (2,736 data points each). Human responses serve as ground truth.}
\label{tab:conf_matrix_after}
\end{table*}

\comment{From these aggregated matrices, we compute precision and recall:
\begin{itemize}
\item \textbf{Before:} Precision = $\frac{49}{49+407}$ = 0.11, Recall = $\frac{49}{49+129}$ = 0.28, F1 = 0.16
\item \textbf{After:} Precision = $\frac{138}{138+318}$ = 0.30, Recall = $\frac{138}{138+111}$ = 0.55, F1 = 0.39
\end{itemize}}

\paragraph{Results Discussion.}
Our first goal was to validate that the model’s predictions align with human pareidolic perception. The results confirm this alignment: the model achieved 0.28 recall before enhancement, significantly outperforming the random baseline (0.07 recall). This demonstrates that our model successfully captures meaningful pareidolic patterns that resonate with human perception.

Building on this foundation, our second goal was to enhance these perceptual alignments. After enhancement, performance improved dramatically across all metrics: recall nearly doubled (from 0.28 to 0.55), precision nearly tripled (from 0.11 to 0.3), and the F1 score more than doubled (from 0.16 to 0.39). In parallel, user behavior reflected similar gains. The proportion of participants who reported seeing zero animals decreased from 54\% to 42\%, while responses shifted toward identifying one or two animals per image, as shown in Figure~\ref{fig:histogram_response_counts}.

These results demonstrate the effectiveness of our enhancement approach. Users perceived significantly more of the animals identified by the model (recall nearly doubled) and exhibited greater confidence in their selections (precision nearly tripled). The enhancement successfully guided human perception toward the intended pareidolic interpretations.

Together, these findings validate our overall approach: the system not only anticipates how humans interpret clouds but also effectively steers their perception through naturalistic visual cues, more than doubling the alignment between model predictions and human perception across all metrics.

Our results cannot be directly compared with existing methods, as the task of predicting and enhancing the perception of pareidolic animals in cloud images is novel. Therefore, we evaluate our method’s performance against both an optimal model and a random baseline.

\paragraph{Comparing against an optimal model.} Ideally, we would aim for an F1 score of 1, which is impossible for our data. For the purpose of analysis, we consider an optimal model whose F1 score is maximized on the survey results. Optimizing for a single participant–segment pair is trivial, as the model could simply mimic the participant’s response. However, due to inter-participant variability in pareidolic perception, perfect agreement with all users is inherently unattainable. Moreover, we evaluate F1 on the aggregated confusion matrix. 
Therefore, we adopt the following heuristic: the optimal model predicts a given animal in a given segment only if at least $T\%$ of participants perceived that animal.
Through an exhaustive search across all threshold values of $T$, we found the optimum at 26\% agreement, achieving an F1 score of 0.45 - comparable to the F1 score of 0.39 achieved by our model (see Table~\ref{tab:baseline_comparison}).


\paragraph{Comparing against a random baseline.}
To verify that our model’s predictions are meaningful, we compare them against a random-model.
For each of the 9 animals and a segment,  we simulate a random-model 
flips a coin (with a 0.5 probability) and use the results as the model prediction. 
We performed the same analysis on the random-model as for our model on the aggregated confusion matrices.

Table~\ref{tab:baseline_comparison} presents the resulting scores. It can be shown that our model substantially outperforms the random predictor. 
Even before enhancement, our model outperforms random chance across all metrics. After enhancement, performance increases dramatically. This confirms that (1) the model’s predictions are meaningfully aligned with human perception, and (2) the enhancement process significantly strengthens this alignment.

\begin{table}[b]
\centering
\begin{tabular}{|l|c|c|c|}
\hline
\textbf{Method} & \textbf{Precision} & \textbf{Recall} & \textbf{F1} \\
\hline
Random Baseline & 0.06 & 0.07 & 0.06 \\
Optimal Model & 0.42 & 0.48 & 0.45 \\
Ours Before Enhancement & 0.11 & 0.28 & 0.16 \\
Ours After Enhancement & 0.30 & 0.55 & 0.39 \\
\hline
\textbf{$\Delta$ (After, Random)} & \textbf{+0.24} & \textbf{+0.48} & \textbf{+0.33} \\
\hline
\end{tabular}
\caption{Comparison of model performance versus random baseline and optimal baseline.}
\label{tab:baseline_comparison}
\end{table}

\subsection{Technical Details}
To efficiently run animal generation on our collected dataset, we used an NVIDIA GeForce GPU with 12GB of VRAM and implemented our method using PyTorch and Hugging Face Diffusers. Each cloud transformation was optimized over 200 steps ($t = 200$), with an average generation time of 60 seconds per image per animal.

For the user survey, we processed the original images by blurring the background outside the mask and highlighting the occluded contour of the segmented cloud. This focused participants’ attention on the relevant cloud segment and minimized distractions from the surrounding sky. Additionally, we generated enhancement videos for `None' segments to help validate the results.

\section{Summary and Future Work}
We introduced a novel framework for predicting and enhancing the perception of pareidolic animals in cloud images. To the best of our knowledge, this is the first computational approach designed to both predict and influence the perception of animals in clouds.

Our zero-shot method leverages a variation of DDS, termed Masked Delta Denoising Score (MDDS), to improve recognizability without requiring model retraining.

A user perceptual study demonstrates that our approach overcomes the limitations of state-of-the-art recognition methods in predicting pareidolia. Moreover, it shows that our enhancement videos significantly improve human perception of pareidolic animals.

Currently, the range of animals our method can detect is constrained by the recognition categories supported by OneFormer, a limitation that could be mitigated by integrating open-vocabulary recognition models.
While our study focuses on cloud imagery, pareidolia also appears in art, astronomical imagery, and abstract natural textures. Our method may thus serve as a foundation for exploring pareidolia in other domains, contributing to future research at the intersection of machine perception and cognitive science.




{\small
\bibliographystyle{ieee_fullname}
\bibliography{references} 
}
\end{document}

%% file: preamble.tex
\newcommand{\red}[1]{{\color{red}#1}}

\newcommand{\TODO}[1]{\textbf{\color{red}[TODO: #1]}}
\renewcommand{\TODO}[1]{}

\usepackage{microtype}

